# Pilot selection in the era of Virtual reality: algorithms for accurate and interpretable machine learning models

Luoma Ke [1], Guangpeng Zhang[2], Jibo He [1,*], Yajing Li[1], Yan Li[1], Xufeng Liu[3], and Peng Fang[3]

[1] Psychology Department, School of Social Sciences, Tsinghua University; klm21@mails.tsinghua.edu.cn
[2] School of Electrical Engineering, Zhengzhou University
[3] Department of Military Medical Psychology, Air Force Medical University

**Abstract:** With the rapid growth of the aviation industry, there is a need for a large number of flight crew. How to select the right pilots in a cost-efficient manner has become an important research question. In the current study, twenty-three pilots were recruited from China Eastern Airlines, and 23 novices were from the community of Tsinghua University. A novel approach incorporating machine learning and virtual reality technology was applied to distinguish features between these participants with different flight skills. Results indicate that SVM with the MIC feature selection method consistently achieved the highest prediction performance on all metrics with an Accuracy of 0.93, an AUC of 0.96, and an F1 of 0.93, which outperforms four other classifier algorithms and two other feature selection methods. From the perspective of feature selection methods, the MIC method can select features with a nonlinear relationship to sampling labels, instead of a simple filter-out. Our new implementation of the SVM + MIC algorithm outperforms all existing pilot selection algorithms and perhaps provides the first implementation based on eye tracking and flight dynamics data. This study's VR simulation platforms and algorithms can be used for pilot selection and training.

## 1. Introduction

For any country, pilots are an indispensable strategic resource for national safety and civil air transportation. Thus, pilot selection is crucial, which can improve the flight performance of the pilot, potentially increase the training success rate and reduce training cost [1]. Furthermore, training cost reduction is an important organizational goal for the aviation industry (i.e., including military and civil aviation), especially during current down seasons of COVID infection. For example, the cost per person who fails to complete undergraduate pilot training is estimated to be $80,000 in the US Air Force [2]. According to a previous study, personal selection impacts training costs and organizational productivity [3]. According to Boeing's 2015-2034 market outlook, 588,000 new commercial airline pilots will be needed worldwide over the next 20 years [4], which imposes an urgent need for better, more accurate, and lower costs for pilot selection methods. In brief, pilot selection is vital for the aviation industry; an appropriate algorithm and platform for pilot selection can reduce training attrition, enhance flight performance, and improve organizational efficiency.

The current pilot selection method still needs improvement to achieve the above goals for pilot selection in at least three aspects. First, pilot selection is conducted by most civil or military pilot training schools or airlines [5, 6]. Traditionally, the pilot selection process uses predictors of performance from cognitive ability tests; however, unfortunately, the moderate effectiveness of these measurements cannot live up to the expectations of pilot selections [7]. A recent study showed only weak correlations between self-report and behavioral measures of the same construct, and it further suggested that these weak correlations were due to the poor reliability of many behavioral measures and the different response processes involved in the two measure types [8].

Second, the performance rating in a flight simulator can predict pilot selection success with mean validities ranging from .24 to .35, overweighing the predictability of the cognitive ability test [9]. Unhesitatingly, these traditional flight simulators have unique advantages for pilot selection over personality questionnaires and cognitive task performances. At the same time, their drawbacks are distinct (e.g., high cost, insufficient immersiveness, environment unfriendly, etc.) [10]. Third, machine learning (ML) modeling (or artificial intelligence) has gradually been applied in personnel selection in recent years [11]. ML algorithms are able to handle a larger number of predictor variables than classical statistical modeling and often reflect complex interactions and nonlinearities, and thus ML models typically exhibit high predictive powers [12]. However, some ML methods are not interpretable and cannot incorporate expert or regulatory knowledge. Interpretability for ML algorithms is expected by the National Institute of Standards and Technology (NIST) and many researchers, and is critical for the fairness, acceptance, and training of the pilot selection [13]; Machine learning algorithms may help to build pilot selection models of a transparent and interpretable decision-making procedure.

Machine learning features for pilot selection should include eye movement metrics as piloting is mostly a visual behavior (see detailed reviews [14, 15]). Many studies have shown that the eye-scanning mode significantly differs between novice and expert [16-18]. Experts' gaze duration on the instrument was shorter and the frequency of gaze on the instrument was higher than that of novices [19, 20]. Although eye movement metrics facilitate aviation research, there is room for improvement in the method based on traditional eye-tracking tools. For example, the GazePoint GP3 desktop eye tracker was used to evaluate the pilot selection process, and the results showed a positive correlation of eye movements with the usual paper and pencil-based selection tests [21]. However, this traditional desktop eye tracker is time-consuming and unfriendly for participants (e.g., participants were asked to keep their heads still through the whole process); on the contrary, Virtual Reality Eye Tracker (e.g., HTC VIVE Pro Eye) can be analyzed with automated analysis scripts without time-consuming manual coding of the areas of interest (AOIs) and friendly for participants (e.g., participants can move around without restrictions). Furthermore, to our best knowledge, the Virtual Reality Eye Tracker is rarely used in pilot selection, if never attempted.

Another type of feature for machine learning is flight dynamics, which is often evaluated qualitatively and crudely by peer interviewees, but seldomly quantitively evaluated and used in machine learning. The quick access recorder (QAR) keeps various pilot operating and aircraft parameters, environmental and alarming information [22]. Which parameters recorded by QAR could be a candidate indicator for pilot selection? The existing literature provided us some information, but not the full scope of the contribution of flight dynamics data to pilot selection. For example, studies demonstrated that the flight performance can be measured as the deviance from the ideal traffic pattern [23, 24], and the pitch angle [25]. Useful as it is, QAR or flight dynamics data are not sufficiently utilized in pilot selection. Algorithmic assessment based on the QAR quantification is more reliable than an instructor looking at the flight path and evaluating flight performance subjectively. No studies use QAR or flight dynamics to select pilots quantitatively by machine learning algorithms after extensive literature search by our research team.

### Algorithms of Pilot Selection

In addition to the aforementioned necessity to use a VR platform and machine learning to select pilots, two core research questions remain to be answered: firstly, what input features are available and should be used for machine learning; secondly, what's the best-performing machine learning algorithm for pilot selection? The following two sections address the two core research questions. See Table 1 for a summary of input features and algorithms for pilot selection.

First, input features for machine learning can potentially include personality, cognitive task performance, electroencephalography (EEG), heart rate, eye movements, and flight dynamics.

1.  More specifically, for personality features, Cattell's 16PF-personality scale using the Support Vector Machine (SVM) generated an accuracy of 64% and 78% in two studies by researchers at The Fourth Military Medical University, China [26, 27]. A highly relevant summary whitepaper titled "The predictive power of assessment for pilot selection" generated by the cut-e Group, a consulting company in Germany specializing in pilot selection reported that a job success prediction accuracy of 79.3% can be achieved using personality characteristics, flight simulator results and prior flying experience [28].

2.  Cognitive tasks are the commonly used predictors in pilot selection. These cognitive tasks include General Mental Ability (such as general ability, verbal ability, quantitative ability or the *g*-factor), spatial ability, gross and fine dexterity, perceptual speed, etc. [29, 30]. Cognitive tasks have the advantages of being low cost and easy to implement with paper and pencils or a computer, compared to EEG, eye movement, and flight dynamics. Two studies with cognitive tasks as subcomponents to select pilots achieved a predictability of accuracy in the range of 74% to up to nearly 94% [31, 32].

3.  EEG. Only one study using EEG and machine learning was identified to select pilots [33]. The rare use of EEG to select pilots is perhaps because of the technical difficulty, intrusiveness, and more than 30 minutes of EEG preparation time to use traditional EEG. The EEG components used in their SVM machine learning classification were the power spectrum factor of alpha, theta, and delta waves at the O1, Oz, and O2 electrodes, and their relative power of the three EEG waves. As summarized in Table 1, a classification accuracy of 76% was achieved for a combination of EEG, heart rate, and eye movement [33], and each component's predictability accuracy is unknown.

4.  Eye movement. Similar to EEG, quite a little research has been done on using eye movement and machine learning to select pilots. Only one study considered three eye movement parameters in selecting pilots: blink rate, average gaze duration, and pupil diameter [33]. Although they achieved an overall 76% prediction accuracy, no independent contribution of eye movement was provided in their work [33]. Despite little work on the utilization of eye movement in the pilot selection, eye movement was able to distinguish novice and expert vehicle drivers [34], and can be used to predict driver cognitive distraction with a high accuracy of 90% using machine learning algorithms like SVM [35]. Recent advances in VR-based eye-tracker can potentially reduce traditional eye-trackers costs and manual coding efforts, which might make the application of eye-trackers in pilot selection more feasible and practicable.

5.  Flight dynamics measured by flight simulator or QAR is often considered in pilot selection [28]. The seminal work published in Psychological Bulletin after reviewing 85 years of research in pilot selection reported that the mean validity of 0.63 can be achieved with a combination of general mental ability and a work sample test [30]. Despite the high predictability of flight dynamics, no published scientific study on pilot selection was identified using flight dynamics and machine learning to our best efforts and knowledge. Flight dynamics are often rated crudely via peer ratings, which generates much lower validity than a work sample test (0.49 compared to 0.54 respectively) or flight dynamics [30].

**Table 1.** Summary of Algorithms and Input Features of Pilot Selection.

| Index | Algorithms | Input Feature | Accuracy | Institute, Country | References |
|-------|------------|---------------|----------|--------------------|------------|
| 1 | Support Vector Machine | Cattell's 16PF-personality | 78% | The Fourth Military Medical University , China | [26] |
| 2 | Support Vector Machine | Cattell's 16PF-personality | 64% | The Fourth Military Medical University , China | [27] |
| 3 | extremely randomized tree | Cognitive task performance & personality test etc. | Nearly 94% | United States Air Force Academy , United States | [31] |
| 4 | discriminant analysis | Cognitive task performance | 74% | ISPA-Instituto Universitário , Portugal | [32] |
| 5 | logistic regression | Cognitive task performance | 77% | ISPA-Instituto Universitário , Portugal | [32] |
| 6 | neural network | Cognitive task performance | 76% | ISPA-Instituto Universitário , Portugal | [32] |
| 7 | Support Vector Machine | EEG, heart rate measured using ECG, eye movements (blink rate, gaze duration, and pupil diameter) | 76% | Beihang University , China | [33] |

Second, what's the machine learning algorithm for pilot selection? As summarized in Table 1, prior researchers have explored mostly Support Vector Machine (SVM), extremely randomized trees, discriminant analysis, logistic regression, and neural network etc.

As SVM was used mostly according to the summary in Table 1, thus, we conceive a machine learning framework using an SVM classifier combining a mutual information coefficient (MIC) feature selection method to identify pilot experts from novices, in which the MIC method is proposed in a hope to compliment and improve the performance of existing SVM algorithm further. SVM as a powerful algorithm for classification or regression tasks has been proven superior to many related algorithms [36]. Training SVM is to solve a quadratic programming problem, essentially. Given a training dataset defined as $D(x_i, y_i)$, $x \in R^d, y \in \{0, 1\}$. The optimal hyperplane determined by $g(x)$ as follows:

$$g(x) = \text{sign}(\phi(w) \cdot \phi(x) + b)$$

where $\phi(w) \cdot \phi(x) = K(w, x)$, $K(\cdot)$ denotes the kennel function, $\text{sign}(\cdot)$ is the symbolic function, and $b$ is a bias. Different optimization methods of kernel function will be different. Then the optimization of SVM is:

$$\phi^*(w) = \underset{\phi(w)}{\text{argmin}} \frac{1}{2} \phi(w)^2$$

$$s.t. \quad y_i(\phi(w) \cdot \phi(x_i) + b) \geq 1, i = 1, 2, \dots n$$

Lagrange multiplier method is used to solve quadratic convex optimization:

$$L(\phi(w), b, \alpha) = \frac{1}{2} \left| \left| \phi(w) \right| \right|^2 - \sum_i \alpha_i(y_i(\phi(w) \cdot \phi(x_i) + b) - 1)$$

where $\alpha$ represents the Lagrangian multiplier. The MIC feature selection ranked the importance of features by estimating the mutual information coefficient between features and classes. It can be described as follows:

$$MIC(X, Y) = H(X) + H(Y) - H(X, Y)$$

which $H$ is an entropy, and $H(X, Y)$ is a joint entropy of inputs of $X$ and $Y$. In order to prove the superior performance of SVM combining MIC than previously ordinary SVM [26, 27], experiments on multiple classifiers and feature selection algorithms were carried out and compared.

To sum up, in order to meet the need for pilot selection and leverage the new VR technology, the current study attempted to use the seldomly utilized eye movement and flight dynamics (QAR-like data) as features for machine learning, and explore multiple machine learning predictors and feature selection methods to identify the best-performing algorithm, with the consideration of interpretability need of the machine learning algorithms.

## 2. Methods

### 2.1. Participants

A total of 46 righthanded participants (all males) completed the experiment (age range from 25 to 58 years, expert mean age: 32.52 years, *SD* = 7.28, novice mean age: 29.57 years, *SD*=5.74). Among all participants, twenty-three expert participants were recruited from China Eastern Airlines, and the other 23 novice participants were from the community of Tsinghua University.

The research protocol in the current study have received approval from the Institutional Review Board (IRB) of Tsinghua University (ethical approval code: 2022 Ethic Audit No.17). All participants signed informed consent forms and were informed that they could drop out the experiment at any time without any forms of penalty. They were all paid 80 Chinese Yuan (about 12 US dollars or a present of equal value) for their participation.

### 2.2.Procedure

#### 2.2.1 Flight Task

The flight task is a traditional traffic pattern, which mainly includes six stages: upwind, crosswind, downwind, turning base, base, and final (see Figure 1). Participants are asked to put on the HTC Eye Pro VR and finish the flight task in the VR flight simulator.

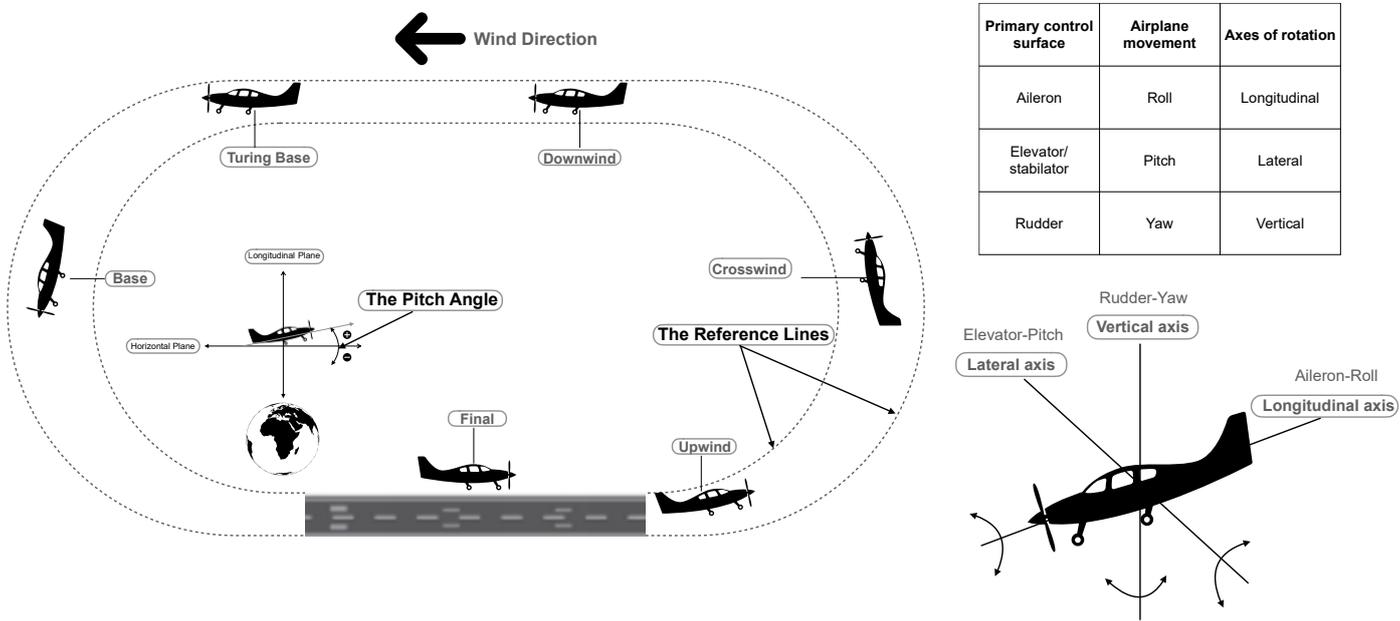

**Figure** 1. The objective measures of the flight performance related to the traffic pattern.

## 2.2.2 Experiment Process

The experiment is made up of four steps, which include "Information collection," "Flight training," "Formal flight," and "Data collation." The experiment flowchart is shown in Figure 2.

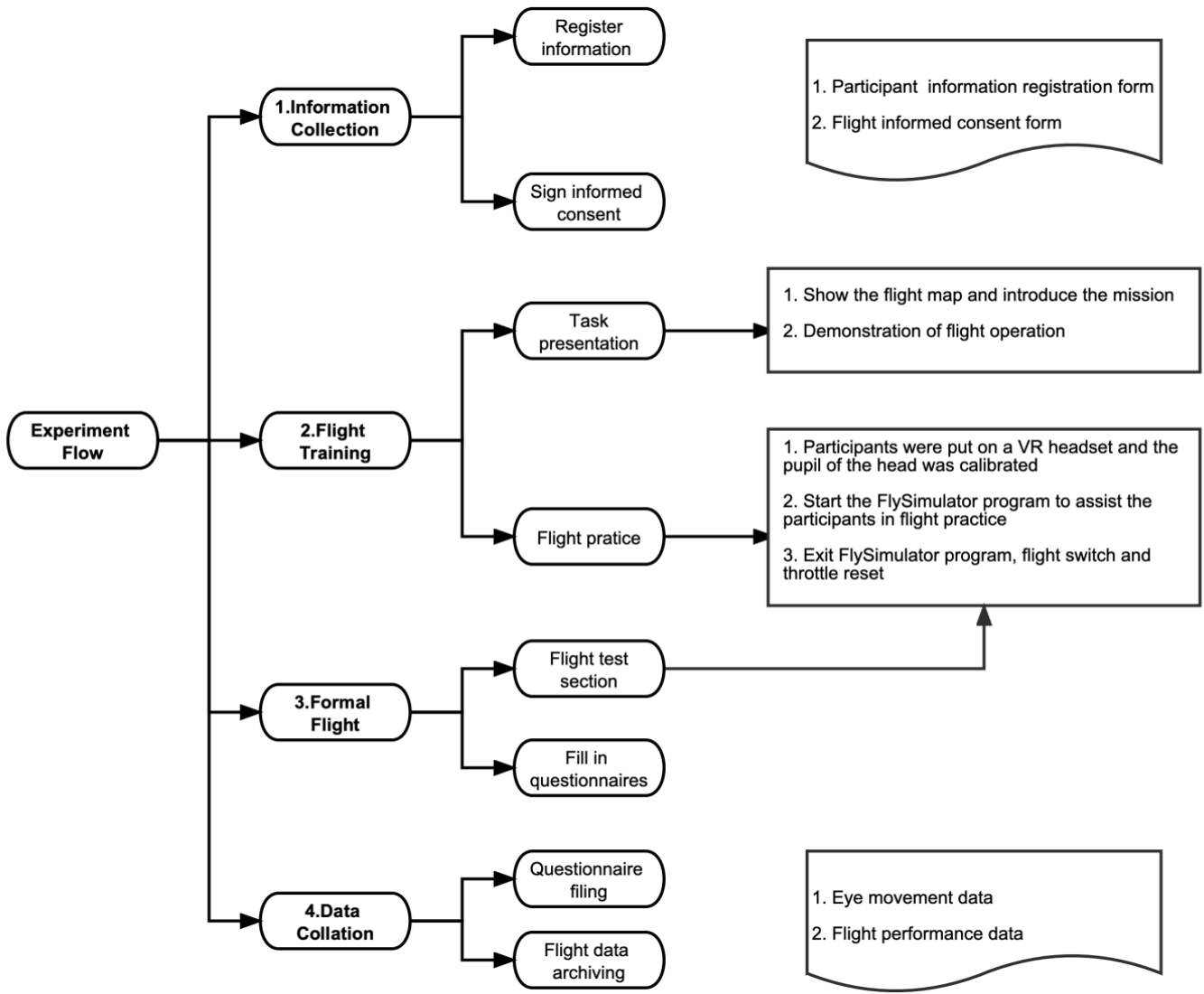

**Figure 2.** Experiment process flowchart.

*2.3.Feature selection method*

In pattern recognition, machine learning, and data mining, feature selection is an effective means of data preprocessing [37]. It typically works by reducing redundancy features of data. They decrease the computation complexity of models, improve the model's robustness, and avoid overfitting problems [38]. In essence, feature selection is an NP-hard problem because it aims to select a subset from $2^N$ possible subsets of the dataset, where $N$ represents the number of features of the dataset [39]. Therefore, it is efficient to use feature selection methods to search for proper combination of features in a polynomial time [40].

Three feature selection algorithms were implemented and compared in this study, namely, mutual information coefficient (MIC), support vector machine-based recursive feature elimination (SVM-RFE), and random forest (RF). According to whether feature selection algorithms use classifiers, feature selection algorithms are divided into three categories, which are filter methods (classifier-independent), wrapper methods (classifier-dependent), and embedded methods (classifier-dependent) [41]. As an instance of filter methods, MIC ranks features by the estimated mutual information. SVM-RFE, an embedded method, selects the relevant features by their default settings. RF, as a wrapper method, orders features by feature importance vector. Combining these feature selection methods, a strategy of selecting a subset of features based on the percentage of ranked features is used.

*2.4. Predictors*

Pilot selection is a binary classification task in this study. Concretely, the binary predictors were fed as an input of the participant's features and output 1 or 0, where 1 means expert, and 0 means novice. In the present study, we applied feature selection techniques to a dataset and subsequently evaluated the performance of five binary classification algorithms, namely, support vector machine (SVM) [42], k-nearest neighbor (KNN) [43], logistic regression (LR) [44], light gradient boosting machine (LGBM) [45], and decision tree (DTree) [46].

*2.5. Cross validation*

Cross validation is a common method to evaluate model performance in machine learning [47]. It divides the training data into several subsets and uses one subset each time to verify the model obtained from the training of other remaining subsets, so as to reduce the error caused by unreasonable partitioning of training data. When the number of instances in a dataset or the number of a certain class is too small, it may be beneficial to use a technique called "leave-one-out" cross validation to obtain a more reliable estimate of the accuracy of a classification algorithm [48]. This is a special condition of k-fold cross validation, where only one instance is left out during each iteration. This allows for the use of all the data in the estimation process, while still providing a way to measure the algorithm's performance on unseen data [49].

*2.6. Metrics*

This study adopted five commonly used metrics to evaluate our algorithm, i.e., F1 score (F1), Accuracy (Acc), area under the receiver operating characteristic (ROC) curve (AUC), Precision, and Recall. AUC calculated by the area under the ROC curve is a widely used indicator to assess an unbalanced learning performance [50]. If an AUC metric must be labeled as good or bad, we can reference the following rule-of-thumb from Hosmer and Lemeshow in Applied Logistic Regression (p. 177): 0.5 = No discrimination, 0.5-0.7 = Poor discrimination, 0.7-0.8 = Acceptable discrimination, 0.8-0.9 = Excellent discrimination, and >0.9 = Outstanding discrimination. By these standards, a model with an AUC score below 0.7 would be considered poor and anything higher than 0.7 would be considered acceptable or better. The confusion matrix is a concept in Psychophysics and machine learning, especially in Signal Detection Theory. The measurement of the confusion matrix includes FP, TP, FN, and TN, where TP and FN are true positive numbers and false negative numbers respectively; TN and FP represent the number of true negative and false positive. The formulas of F1, Acc, AUC, Precision, and Recall are as follows:

$$F1 = \frac{2 \times Precision \times Recall}{Precision + Recall}$$

$$Acc = \frac{TP + TN}{TP + TN + FP + FN}$$

$$AUC = \frac{\sum_{i=1}^{n^+} \sum_{j=1}^{n^-} 1_{if(p^+ > p^-)}}{n^+ n^-}$$

$$Precision = \frac{TP}{TP + FP}$$

$$Recall = \frac{TN}{TP + FN}$$

where $n^+$ is the number of positive samples, $n^-$ is the number of negative samples, $1_{if(p^+>p^-)}$ represents 1 when $p^+ > p^-$, and $p^+$ and $p^-$ are the prediction probabilities of positive and negative samples.

*2.7.Data analysis*

2.7.1 Flight performance data

Four major indicators were applied to evaluate the flight performance in the current study: The total flight time (i.e., time from takeoff to landing, in seconds), pitch angle 1s before landing (i.e., the pitch angle of the plane 1 second before landing, in degrees), mean distance to the center of the two reference lines (in meters), and standard deviation of distance to the center of the two reference lines (in meters).

2.7.2 Eye movement data

Four key flying instruments (i.e., airspeed indicator, vertical speed indicator, attitude indicator, and altitude indicator) were defined as different areas of interest (AOIs) in the study. Each AOI's percent dwell time (unit: %) was used to evaluate visual attention dispersion. The mean percent dwell time was the cumulative time observed within the AOI divided by the total gaze time and averaged across participants.

2.7.3 Statistic analysis

Statistical analyses were performed using SPSS Version 24.0 (IBM Corp, Armonk, NY, USA). The significance level was set at $p < 0.05$. The scientific graphs with statistical results were plotted using GraphPad Prism version 9.4.0 for Mac Operating System(GraphPad Software, San Diego, California, USA).

2.7.4 Eye movement preprocessing & analysis

The raw data provided by the HTC Eye Pro and Tobii SDK and logged by SRanipal Eye Framework include a. timestamp in second; b. Gaze origins in millimeters in X, Y, Z axis for left and right eyes (FOL_X, FOL_Y, FOL_Z, FOR_X, FOR_Y, FOR_Z); c. gaze direction normalized to between -1 and 1 in X, Y, Z axis for left and right eyes ( FVL_X, FVL_Y, FVL_Z, FVR_X, FVR_Y, FVR_Z); d. Eye-opening of left and right eyes (EOL and EOR ); e. pupil position normalized to between -1 and 1 in X and Y axis for left and right eyes (PPLX, PPLY, PPRX, PPRY); f. AOIName logs the 19 predefined areas of interest (AOIs), which include aircraft clocks, airspeed indicator, attitude indicator, vertical speed indicator, radio compass, inlet pressure gauge, altitude indicator, turn-and-slip indicator, aircraft tri-use meter, magnetic course correction calculator, tachometer, cylinder head thermometer, air inlet temperature indicator, current, and voltage meters, tank pressure gauge, spare magnetic compass, left aircraft cockpit glass, front aircraft cockpit glass, and right aircraft cockpit glass.

**Table 2.** Features from eye movements.

| Index | Feature/variable names | Calculation method | Performance indication |
|---|---|---|---|
| 1 | Standard deviation of fixation in x axis. | The standard deviation of FVL_X, e.g., numpy.std(data['FVL_X']) | Indicates the horizontal dispersion of fixations, or how wide participants looked |
| 2 | Standard deviation of fixation in y axis. | The standard deviation of FVL_Y,e.g., numpy.std(data['FVL_Y']) | Indicates the vertical dispersion of fixations, or whether participants looked up and down. This often relates to whether pilots are able to look forward and near areas to guide their flight path. |
| | Standard deviation of fixation in z axis. | The standard deviation of FVL_Z, e.g., numpy.std(data['FVL_Z']) | Indicates the depth of visual attention |
| 4 | Eye opening (%, from 0 to 1) | The mean of EOL or EOR, e.g., numpy.average(data['EOL']) | Indicate how wide the eye opens, which is related to participants' interests and workload. |
| 5 | Percent dwell time (%) on each AOI; | $\frac{AOIName_i}{\sum_{k=1}^{N} AOIName_k}$ (AOIName$_i$ is a specific AOI, such as the altitude indicator; k = 1, 2, … , refer to all the indicators, where N = 19, as we have a total of 19 AOIs. ) | Indicates the relative attention to a specific AOI, reflecting cognitive processing, understanding of information in that AOI. Note: This is labelled as "AOI" information in machine learning section; all others in this table labelled as "EM" (eye movement) |
| 6 | Frequency of AOI transitions (Hz) | The number of sequential pairs (AOIName$_i$, AOIName$_j$) where i!=j , divided by time | Indicates how actively participants look for information from gauges, which suggests understanding meaning of gauges. |
| 7 | Fixation duration (ms) | Use i-VT algorithm to detect fixation, with threshold for velocity of 30o/s | larger value indicates more time spent on processing visual information. |
| 8 | Fixation count | The number of fixations | Indicates actively looking for information. |

2.7.5 Flight dynamics preprocessing & analysis

The flight dynamics data were automatically recorded by the VR simulation software. The raw variable includes (a). timestamp in second; (b). the gesture of the airplane measured using Roll, Pitch, and Yaw (Heading); (c). location of the airplane, including longitude, latitude, above ground level (AGL), and above sea level (ASL); (d). flight movement measurements, including velocity and angle of attack (AoA); (e). control device inputs, including rudder, elevator, and roll inputs; and (f). the (longitude, latitude, and height) of the nearest point on the center of the two reference lines.

The above raw values were converted to features in Table 3 according to the QAR (Quick Access Recorder) analysis method inspired by the 35 BAE-146 aircraft QAR data provided by the National Aeronautics and Space (NASA).

**Table 3.** Features from flight dynamics (QAR data).

| Index | Feature/variable names | Performance indication |
|---|---|---|
| 1 | ldg_time (s) | landing time, that is, the time when the airplane lands, which is used as reference time point for the 1s and 8s before landing. |
| 2 | Vert_accel_landing | Vertical acceleration when landing |
| 3 | AOA (1s or 8s before landing) (degree) | Angle of Attack 1s or 8s before landing |
| 4 | AOA (min & max values) | The minimum and maximum values of Angle of Attack. |
| 5 | Pitch_angle(1s or 8s before landing) (degree) | Pitch angle 1s or 8s before landing |
| 6 | RudderInput (1s or 8s before landing) | Rudder input 1s or 8s before landing. |
| 7 | ElevatorInput (1s or 8s before landing) | Elevator input 1s or 8s before landing. |
| 8 | RollInput(1s or 8s before landing) | Roll input 1s or 8s before landing. |
| 9 | TAS (1s or 8s before landing) ( m/s) | True air speed (TAS) 1s or 8s before landing in unit of m/s |
| 10 | GS(1s or 8s before landing) (m/s) | Ground speed (GS) 1s or 8s before landing in unit of m/s |
| 11 | Velocity_Descent_mean (m/s) | Average descent velocity when landing in unit of m/s |

| 12 | Longitude_err (*mean* + *SD*) (m) | The mean and standard deviation (SD) of the airplane position in the longitude axis relative to the nearest center of the two reference lines. |
|----|-----------------------------------|------------------------------------------------------------------------------------------------------------------------------------------------|
| 13 | Latitude_err (*mean* + *SD*) (m) | The mean and standard deviation (SD) of the airplane position in the latitude axis relative to the nearest center of the two reference lines. |
| 14 | Height_err (*mean* + *SD*) (m) | The mean and standard deviation (SD) of the airplane position in the height axis relative to the nearest center of the two reference lines. |
| 15 | dist_err (*mean* + *SD*) (m) | The mean and standard deviation (SD) of the distance of the airplane position to the nearest center of the two reference lines. |
| 16 | rou (min & max values) | The minimum and maximum values of the turning curvature of the airplane, with larger values indicate possible unsafe sharp turning. |
| 17 | acc_h_max (m/s2) | The maximum value of the vertical acceleration. |
| 18 | acc_xy_max (m/s2) | The maximum values of the acceleration in the horizontal plane. The recommended acc_xy_max value for civil aviation pilots is 1G. |
| 19 | Roll (min & max values) | The minimum and maximum values of Roll angle. |
| 20 | Pitch (min & max values) | The minimum and maximum values of Pitch angle. |
| 21 | slide_length (m) | The distance the airplane travelled after landing until full stop. The upper limit for this value is often 1800m for most airports. |

### 2.7.6 Machine learning modeling

After preprocessing, 19, 7, and 39 features exist in the AOI, EM, and QAR datasets. In Table 3, most rows contain two to three features, such as mean and SD, which adds up to 39 QAR features. This study combined three kinds of data (AOI, EM, and QAR) and then obtained seven combinations of the whole datasets: AOI, EM, QAR, AOI & EM, AOI & QAR, EM & QAR, AOI & EM & QAR. Those datasets all contain 45 participants. One participant was excluded from the machine learning analysis as he was the only instructor pilot whose expertise and age differed from other pilots. To illustrate the predictability of the newly proposed framework based on the SVM + MIC algorithm, comparative experiments were conducted based on the AOI & EM & QAR datasets with other algorithms. The whole machine learning algorithms include three processes: feature selection, training predictors, and evaluating predictors. First, the feature selection methods were used to filter redundant features. Feature selection proportion was set to 15% to 95% with a stride of 10%, which means that a certain proportion of relevant features were selected from the ranked features by the feature selection methods. For example, there are 70 rated features after feature selection in a dataset, then the top 7 features were selected with a feature selection proportion of 10%. Second, training predictors. Selected features were fed into multiple predictors, and a leave-one-out cross validation strategy was applied. Last, evaluating predictors. All algorithms were implemented by Python 3.9.12 software. The hyper-parameters of predictors were default settings in the Scikit-learn 1.2.0 package and not adjusted. Additionally, ablation experiments were conducted on the seven datasets to explore the importance of different data sources.

### 3. Results

### 3.1. Flight performance results

An independent-sample *t*-test is performed on the flight QAR indicators. Results are shown in Table 4 and Figure 3. The difference between novice and expert is significant in the pitch angle 1s before landing ($t(44) = 2.09$, $p < 0.05$, Cohen's $d = 0.62$), the mean distance to the center of the two reference lines ($t(44) = 3.96$, $p < 0.01$, Cohen's $d = 1.17$) and the standard deviation of distance to the center of the two reference lines ($t(44) = 3.53$, $p < 0.01$, Cohen's $d = 1.04$). Experts accomplished the whole flight task numerically quicker than novices, as the total flight time is marginally significant ($t(44) = 1.84$, $p = 0.07$, Cohen's $d = 0.54$).

**Table 4.** An independent-sample t-test results of the flight performance data.

| Indicators | Novice | Expert | *t* | *p* | Cohen's *d* |
| --- | --- | --- | --- | --- | --- |
| | *Mean (SD)* | *Mean (SD)* | | | |
| The total flight time (s) | 902.32 (336.73) | 759.06 (163.58) | 1.84 | 0.07 | 0.54 |
| Pitch angle 1s before landing (°) | -12.54 (29.03) | 3.97 (24.43) | 2.09 | * | 0.62 |
| Mean distance to center of reference lines (m) | 873.89 (818.43) | 176.67 (205.52) | 3.96 | ** | 1.17 |
| The standard deviation of distance to center of reference lines (m) | 675.78 (589.07) | 211.52 (225.76) | 3.53 | ** | 1.04 |

*Note*：* represents $p < 0.05$, ** represents $p < 0.01$.

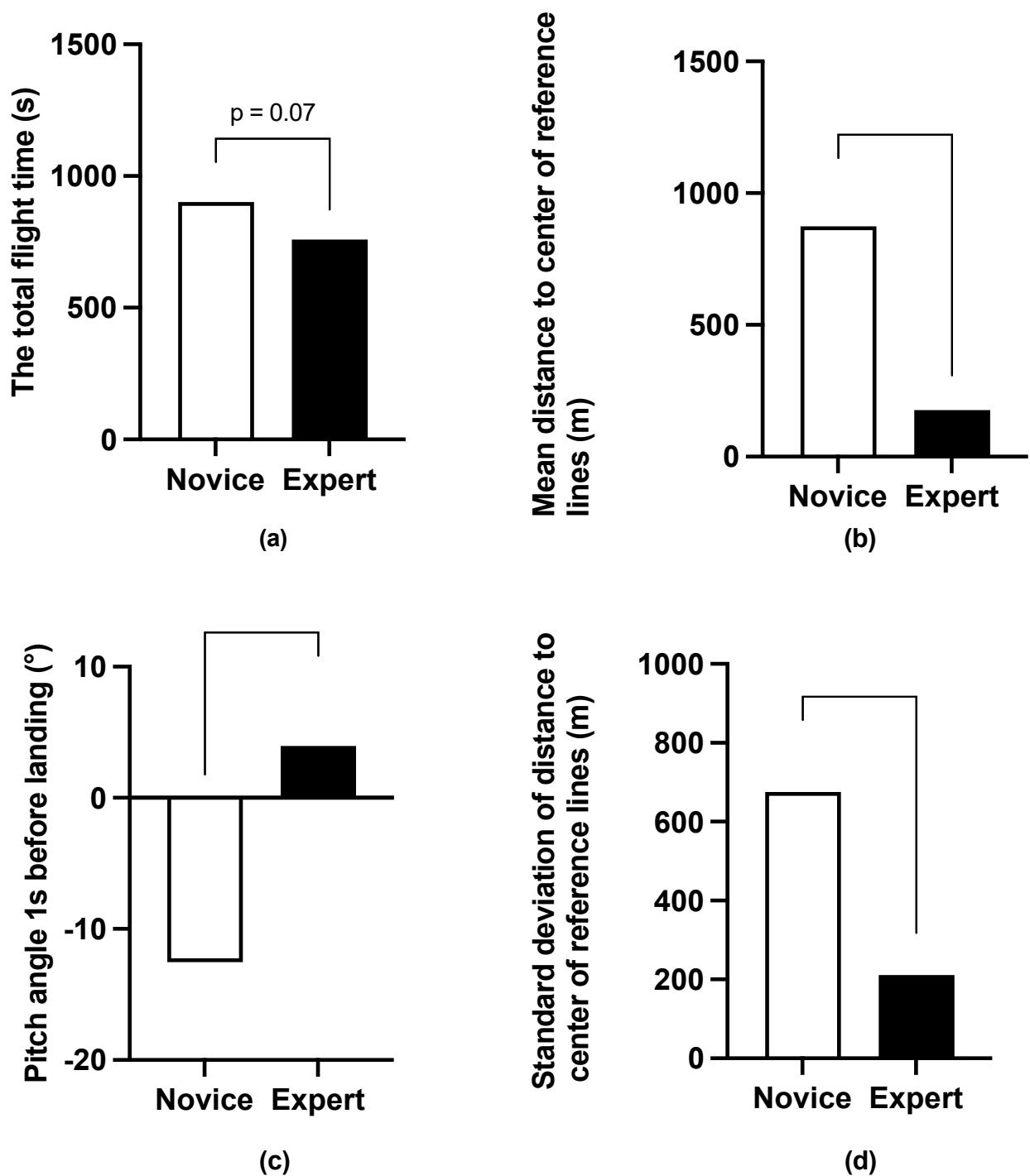

Figure 3. Comparisons of flight performance for novices and experts.

Note: Error bar represents SEM, * represents p < 0.05, ** represents p < 0.01, and *** represents p < 0.001.

### 3.2. Eye movement analysis results

As shown in Table 5 and Figure 4, experts spend a significantly higher percentage of time than novices on the key flying instruments (i.e., airspeed indicator, vertical speed indicator, and altitude indicator): The airspeed indicator ($t(44) = 2.11$, $p < 0.05$, Cohen's $d =$ 0.64), the vertical speed indicator ($t(44) = 3.83$, $p < 0.001$, Cohen's $d = 1.15$), and the altitude indicator ($t(44) = 2.24$, $p < 0.05$, Cohen's $d = 0.68$). However, the difference among experts and novices is only marginally significant in the attitude indicator ($t(44) = 1.84$, $p = 0.07$, Cohen's $d = 0.56$).

**Table 5.** An independent-sample $t$-test results of the percent dwell time on each AOI (Unit: %).

| Area of Interest (AOI) | Novice | Expert | $t$ | $p$ | Cohen's $d$ |
|---|---|---|---|---|---|
| | Mean (SD) | Mean (SD) | | | |
| Airspeed indicator | 5.32 (4.98) | 8.56 (5.45) | 2.11 | * | 0.64 |
| Attitude indicator | 31.03 (12.2) | 25.15 (9.23) | 1.84 | 0.07 | 0.56 |
| Vertical speed indicator | 5.33 (4.11) | 13.11 (8.84) | 3.83 | *** | 1.15 |
| Altitude indicator | 1.34 (2.22) | 2.83 (2.29) | 2.24 | * | 0.68 |

*Note*：* represents $p < 0.05$, ** represents $p < 0.01$, and *** represents $p < 0.001$.

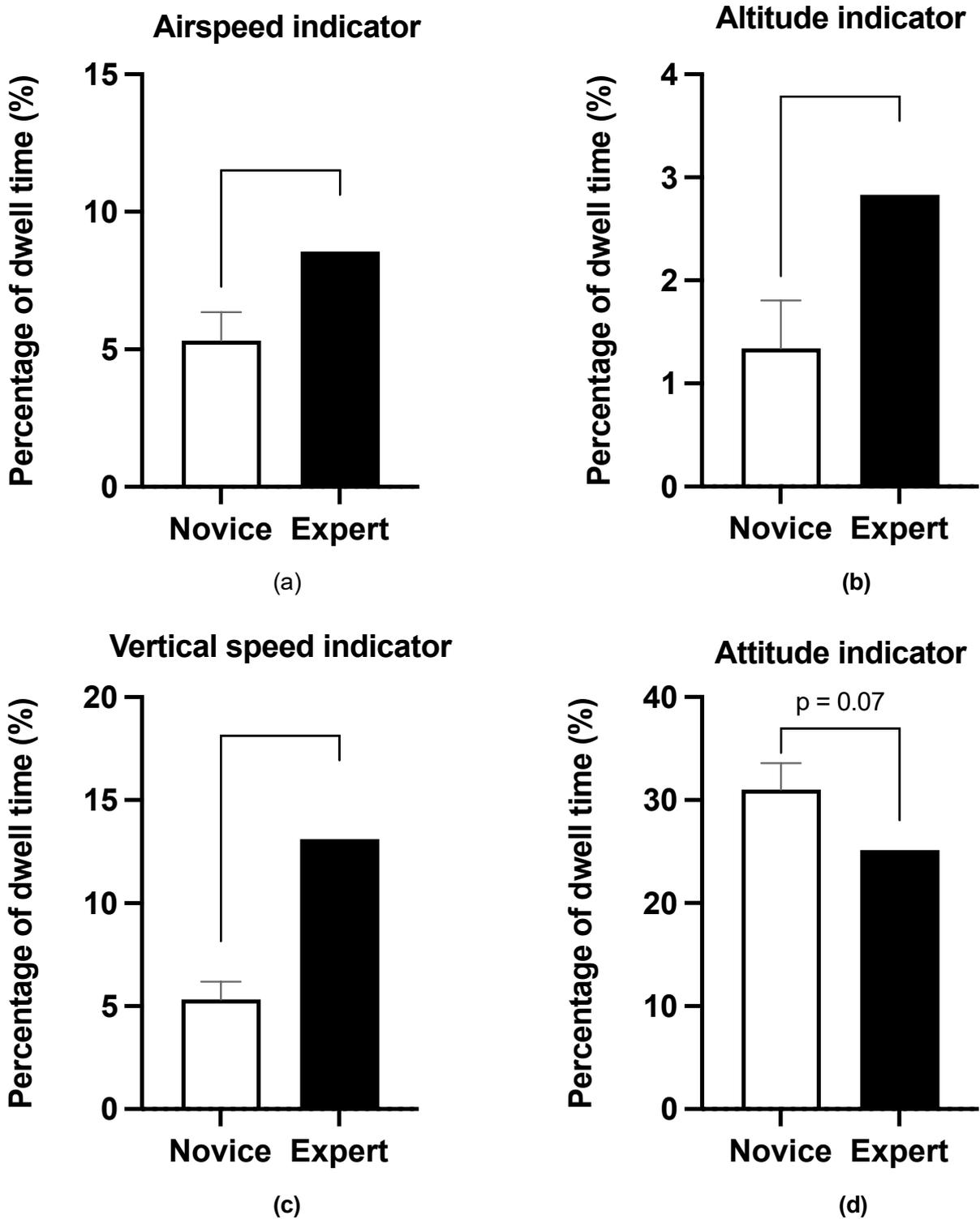

**Figure 4.** Means and data distribution of percent dwell time: (a) Airspeed indicator; (b) Altitude indicator; (c) Vertical speed indicator; (d) Attitude indicator.

Note: Error bar represents SEM, * represents $p < 0.05$, ** represents $p < 0.01$, and *** represents $p < 0.001$.

### 3.3. Evaluation of different proportions of selected features

The strategy of selecting features according to the k highest scores was adopted during the feature selection process, where k denotes an integral part of the product of the feature selection proportion and the number of overall features. Due to the continuity of feature proportions, we cannot exhaustively enumerate all proportions. So, the proportion was set to 15% to 95% with a stride of 10% for evaluating how the prediction performance was influenced. The experiments were conducted on the dataset of AOI & EM & QAR by using nine feature proportions (from 15% to 95% with a stride of 10%), five predictors (SVM, KNN, LR, LGBM, and DTree), and three feature selection methods (MIC, SVM-RFE, and RF) above. Figure 5 shows each proportion's best Acc, AUC, F1, Precision, and Recall performance. 65% feature selection proportion outperformed the other eight proportions on Acc, F1, AUC, Precision, and Recall metrics. Thus, this study adopted the feature proportion of 65% as the best choice.

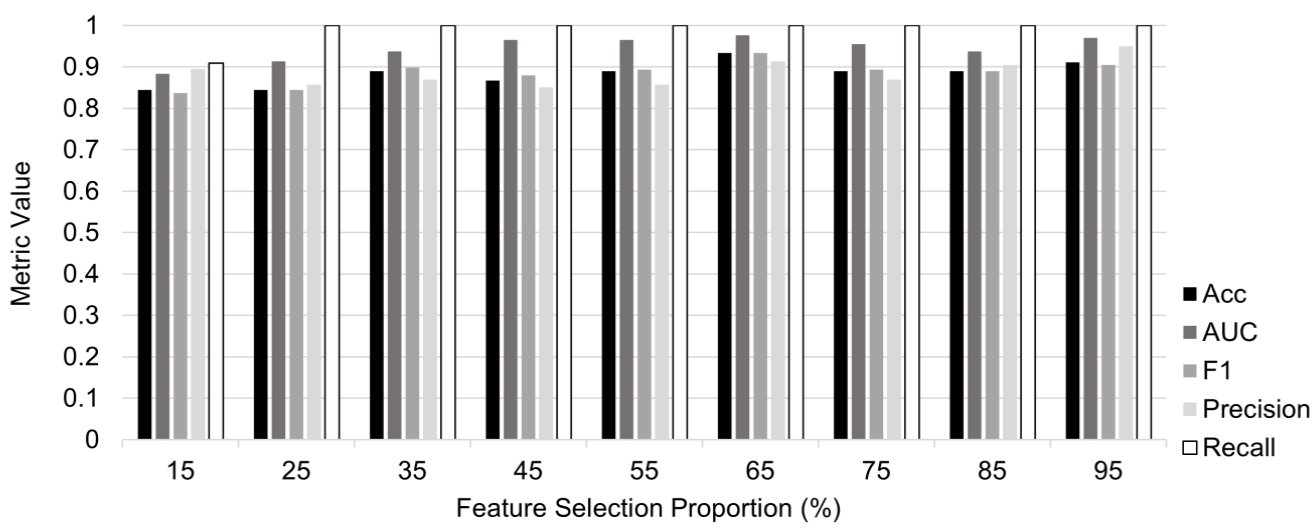

**Figure 5.** Evaluation of different proportions of the features selected on AOI & EM & QAR datasets.

### 3.4. Performance evaluation of predictors and feature selection methods

This section evaluated 15 models with the combinations of three feature selection methods (MIC, SVM-RFE, and RF) and five predictors (SVM, KNN, LR, LGBM, and DTree), as shown in Figure 6. The results indicate that an SVM predictor with the MIC feature selection method generally achieved the highest performance on all metrics with an Acc of 0.9333, an AUC of 0.9644, and an F1 of 0.9333. From the perspective of feature selection methods, the MIC algorithm is the most suitable feature selection method for the pilot selection prediction task. Additionally, SVM performed better robustness in Acc, AUC, and F1 metrics and DTree with an Acc of 0.8889, an AUC of 0.8893, and an F1 of 0.8889 when using the MIC feature selection method. SVM and DTree both showed outstanding discrimination for pilot selection.

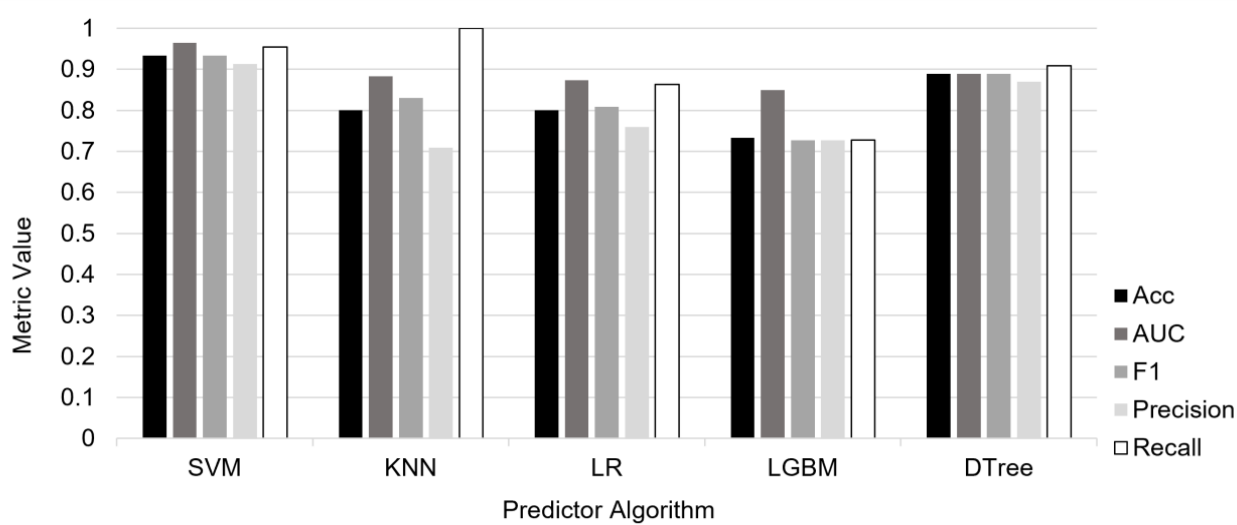

(a). Feature Selection Method: MIC

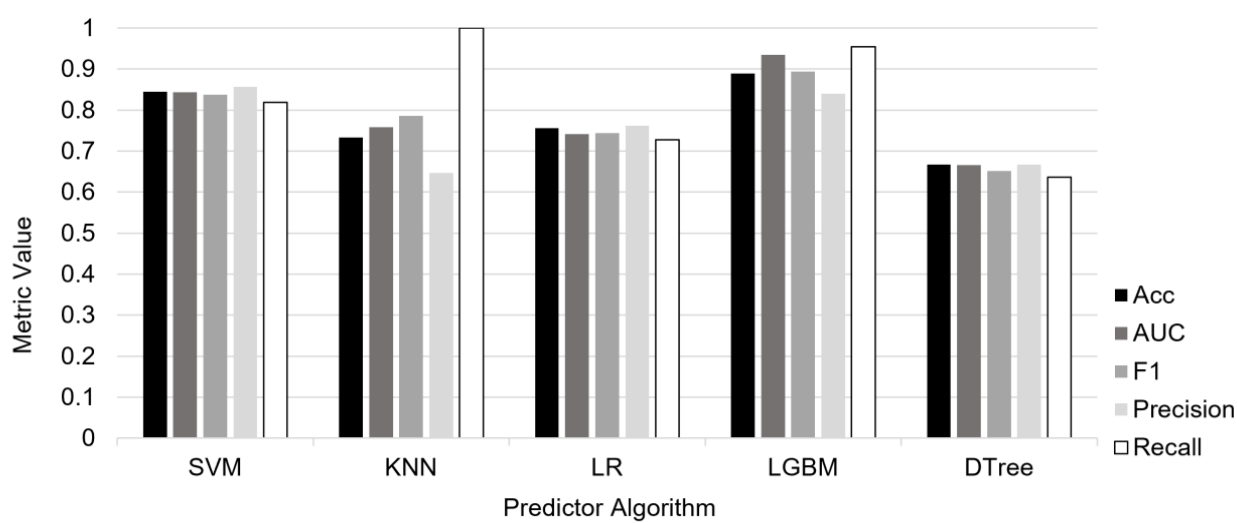

(b). Feature Selection Method: SVM-RFE

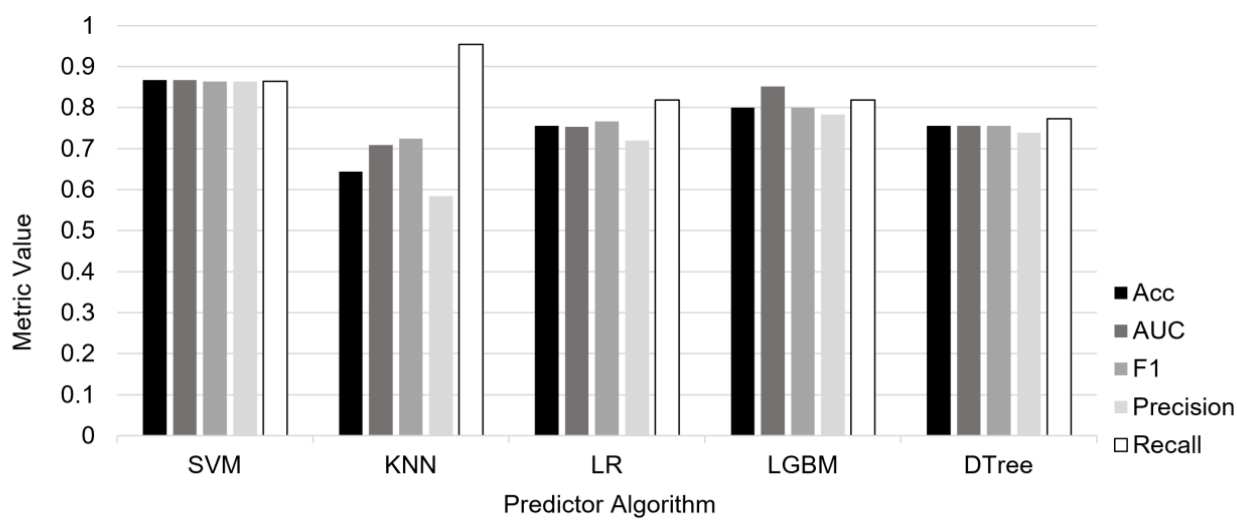

(c). Feature Selection Method: RF

**Figure 6.** The performance comparison between feature selection methods and predictor algorithm.

### 3.5. Ablation experiments on datasets

To investigate the impact of different data sources on the performance of pilot selection prediction tasks. Ablation experiments on the seven datasets were conducted using the SVM+MIC algorithm, which has been proven as the best algorithm for pilot selection as described in the above paragraphs. As Figure 7 shows, the EM & QAR and AOI & EM & QAR datasets achieved AUC values of 96.64% and 96.44%, respectively, and the AOI dataset got the worst AUC value of 82.41%. Table 6 summarizes the prediction performance on the seven datasets. The AOI & EM & QAR dataset obtained the best performance with an Acc of 0.9333, an F1 of 0.933, a Precision of 0.9130, and a Recall of 0.9545. The AOI dataset obtained the worst performance on all five metrics.

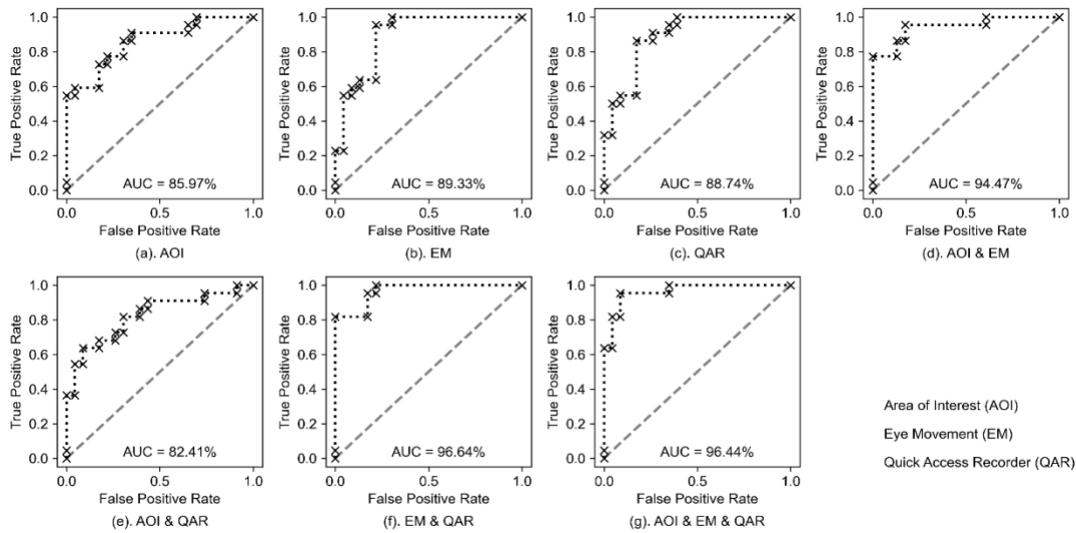

**Figure 7.** The ROC curves of the SVM+MIC algorithm on the seven datasets are (a). AOI, (b). EM, QAR, (c). AOI & EM, (d). AOI & QAR, (e). EM & QAR, and (f). AOI & EM & QAR.

**Table 6.** Prediction performance on ablation experiments for the SVM+MIC algorithm.

| Dataset | Acc | AUC | F1 | Precision | Recall |
|---|---|---|---|---|---|
| AOI | 0.7333 | 0.8597 | 0.7000 | 0.7778 | 0.6364 |
| EM | 0.8222 | 0.8933 | 0.8400 | 0.7500 | 0.9545 |
| QAR | 0.8444 | 0.8874 | 0.8444 | 0.8261 | 0.8636 |
| AOI & EM | 0.8667 | 0.9447 | 0.8696 | 0.8333 | 0.9091 |
| AOI & QAR | 0.7556 | 0.8241 | 0.7317 | 0.7895 | 0.6818 |
| EM & QAR | 0.8667 | 0.9664 | 0.8696 | 0.8333 | 0.9091 |
| AOI & EM & QAR | 0.9333 | 0.9644 | 0.9333 | 0.9130 | 0.9545 |

### 3.6. Interpretable Model Results based Decision Tree (DTree)

Most of the above-mentioned models (Including SVM, KNN, and LGBM) are not interpretable. Although the SVM+MIC algorithm generates the best accuracy of 0.9333, the DTree model is still further analyzed to provide an interpretable model and results for the difference between novice and expert pilots. The DTree with an Acc of 0.8889, an AUC of 0.8893, and an F1 of 0.8889 when using the MIC feature selection method. Figure 8 shows the visualization results of DTree. Thus, when the interpretability of models is emphasized, we need to switch from best performing SVM+MIC algorithm to DTree; we need to trade 0.9333-0.8889 = 0.0444 accuracy for interpretability.

The variables included in the final DTree model include: 1. Percent Dwell Time on Altitude Indicator; 2. Rudder Input 8s before landing; 3. Ground Speed 1s before landing; 4. Elevator Input 8s before landing, and 5. Saccade Count. Figure 9 further depicts the relative contributions of these five variables to the DTree model. Percent Dwell Time on Altitude Indicator contributed most to the DTree model, followed by Ground Speed 1s before landing, Rudder Input 8s before landing, next Saccade Count, and Elevator Input 8s before landing.

An interpretable model thus is that most expert pilots are who:1. Use the Altitude Indicator frequently (larger than 0.013 of the total time); 2. can maintain Ground Speed 1s before landing; 3. Have smaller than 0.4 elevator inputs 8s before landing; 4. Have more Saccade Counts. On the other side, most novices are who:1. Use the Altitude Indicator less frequently, and 2. have smaller Rudder Input 8s before landing.

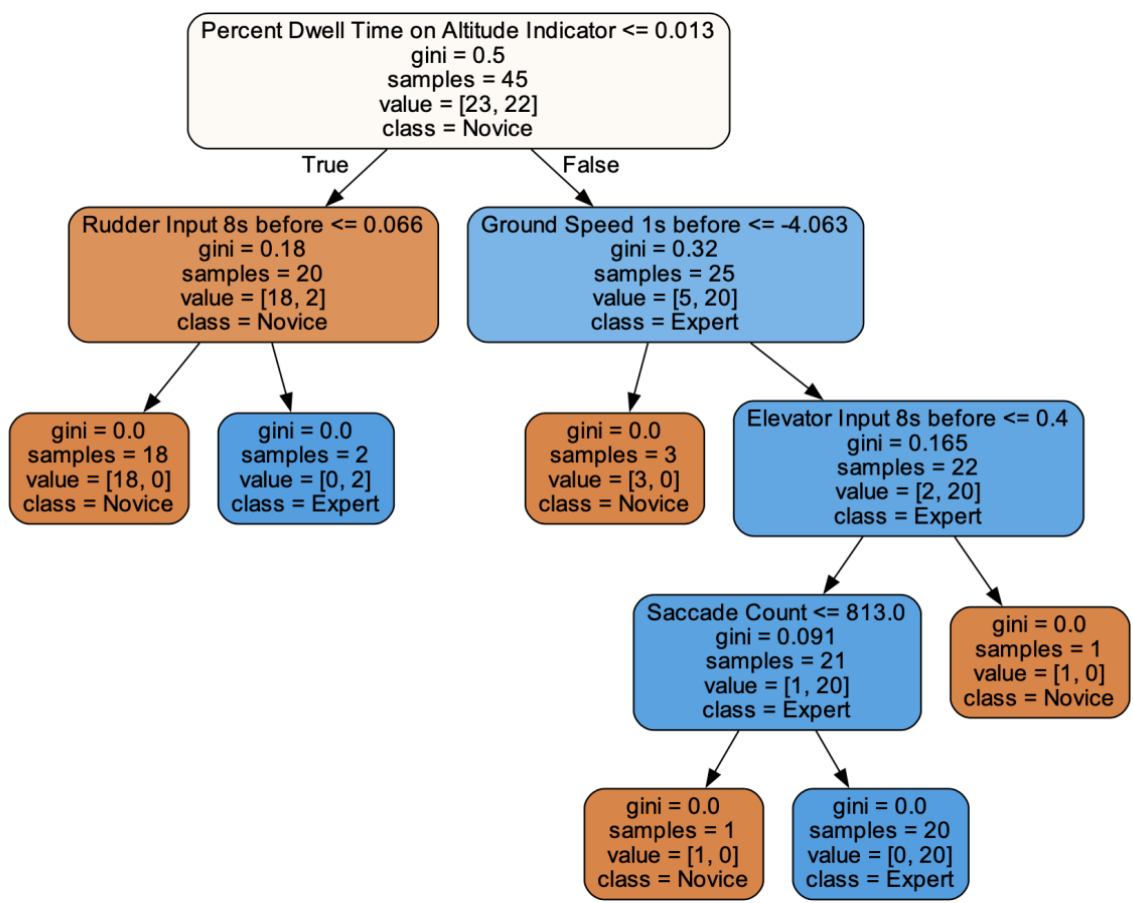

Figure 8. Visualization of Decision Tree classification results.

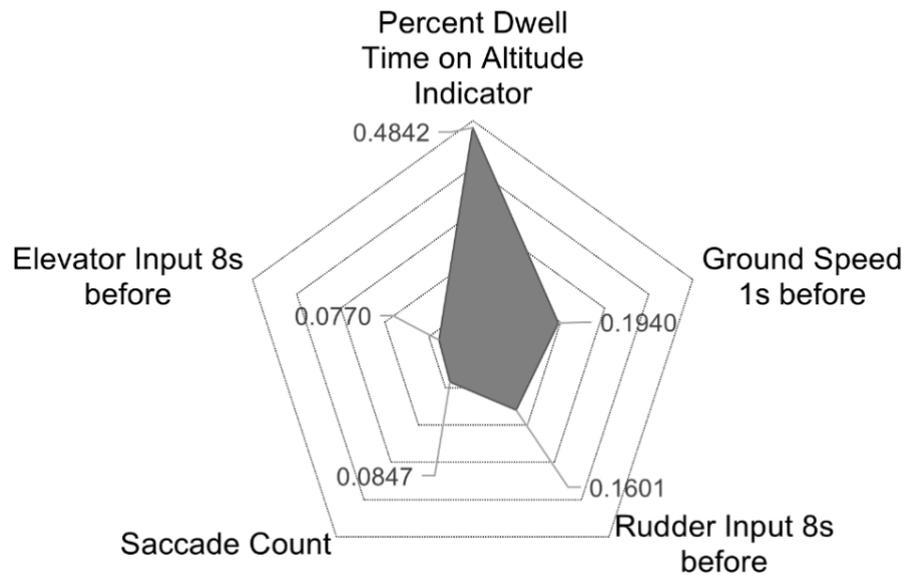

**Figure 9.** Relative Contribution of Variables in the Decision Tree model.

## 4. Discussion

To sum up, for the behavioral results, expert pilots differed from novices in at least three aspects. Firstly, for flight dynamics, the pilots' actual flight path was nearer to the center of the two reference lines than novices. Secondly, pilots had a distinctly different eye-moving pattern from novices. During flying, pilots relied heavily on the key instrumentations of flight, such as the airspeed indicator, the vertical speed indicator, and the altitude indicator. Thirdly, pilots had a longer fixation time on the instruments and exhibited a more structured and efficient pattern of eye scanning. The eye movement results provided new evidence that experts have more efficient eye-tracking models [20].

For the machine learning results, our new SVM+MIC algorithm achieved a high classification of 93.33% in pilot selection. Decision functions of SVM were formed by the specific training dataset. Put another way, SVM can maximize the margin between the decision borders from the dataset in a Euclidean space, making SVM more generalizable, obtaining better robustness, and produce less train error, especially when using a small dataset than other predictors [52]. Moreover, SVM can maximally mine the data's latent knowledge, making small sample prediction tasks possible [53]. One of the most advantages of SVM is fitting nonlinear and high dimensional data better when using nonlinear kernel functions. Mutual information can measure various kind of relationship between random variables, including linear or nonlinear relationships [54]. In addition, Redundant and irrelevant features as data noise decline the performance of the SVM classifier. These are two main reasons why our combined SVM + MIC algorithm achieved the best performances among all evaluated models [55].

The contribution of input features to the predictability of pilot selection was explored by feeding each input features to machine learning algorithms independently or in various combinations. Area of Interest (AOI) dependent analysis was separately analyzed from other measurements of eye movements (labeled as "EM" in Figure 7, including fixation duration, fixation dispersion, and saccade frequency, etc.), as AOI analysis using traditional glass-like eye-tracker and desktop eye tracker often needs time-consuming manual creation and coding of AOIs. As shown in Figure 7 and Table 6, information on AOI-related measurements contributed to an additional 0.9333-0.8667= 0.0666 accuracy gain by comparing "QAR & AOI & EM vs. QAR & EM." To retain this accuracy gained from AOI information, it is important to use a VR-based eye tracker like HTC Eye Pro, instead of a traditional glass-like or desktop eye tracker. In a VR scenario, all objects are in digital forms, thus it is easy and accurate to know the location and semantic meaning of an object or AOI. But traditional eye trackers can only tell researchers where users are looking but cannot tell "what" users are looking at, as current computer vision algorithms cannot recognize all objects' names in videos. In addition, a VR eye tracker (around $1,800) costs way less than traditional eye trackers (around $20,000 to $30,000). Thus, considering time and financial costs, large scales of pilot selection should use a VR eye tracker, which was most likely first implemented and investigated in the current study, instead of traditional eye trackers like GazePoint [21] or SMI ETG Spectacle eye tracker [33]. The time cost and labor to conduct an AOI analysis were perhaps one of the reasons the two studies only conducted a generic common eye movement analysis without AOI information [21, 33]. Nevertheless, the two studies were the only pioneer former studies we have successfully identified which attempted to use eye movement to select pilots. More research efforts are needed for future studies to explore better algorithms to select pilots using eye movement, especially using AOI information and a VR eye tracker.

Existing flight simulator owners may need other information features as inputs for machine learning, such as QAR flight dynamics, as eye trackers are expensive, and more importantly impracticable to set a space for a VR eye tracker in their over 100,000,000 USD advanced flight simulator with 6 degrees of freedom (DoF). Our current work provides a good alternative: QAR flight dynamics for enterprise-level users who cannot choose an eye tracker. As shown in Figure 7 and Table 6, QAR outperforms generic eye movements (see "EM" tick label) and AOI information (see "AOI" tick label). QAR alone can predict pilots with an accuracy of 0.8444, as high as most of the accuracy achievements summarized in Tables 1 and 6.

Although the combination of eye tracking and flight dynamics generates promising discriminability of novices from expert pilots, and these two measurements should be the core of flight behaviors and performances, they are not the complete descriptors of candidate pilots under examination. Other modality variables, such as heart rate, skin conductance, and dry EEG sensors, should be implemented and evaluated in future studies for their predictability of pilot selection success, to provide a platform for pilot selection with all-around modalities.

We are fortunate to locate at least 6 publications that use machine learning to predict pilot selection success, encouraging our work towards the multimodality +machine-learning approach. Especially, for example, a recent publication in 2021 by the United States Air Force Academy claimed that an "*extremely randomized tree machine learning technique can achieve nearly 94% accuracy in predicting candidate success*" based on 8-year data from the historical specialized undergraduate pilot training (SUPT) program [31]. Their long-term work is possibly a milestone that summarizes personnel selection using paper-and-pencils and purely cognitive task performances [31] and signifies the important value of machine learning algorithms in pilot selection. Researchers from the United States Air Force Academy can never be expected and constrained unless the constraint is from historical and technological perspectives. Features with merely cognitive tasks face the limitation of ecological validity and practice effects. Recent advances in Virtual Reality technology with embedded eye-tracking modules can remove the technological constraint for our pioneer researchers [31]. The current work follows their machine learning approach, but with the adoption of new technology, and achieves similar predictability of nearly 94% using multimodality data of eye tracking and flight dynamics.

However, unfortunately, we cannot find papers using machine learning algorithms to select pilots based on data like heart rate [33] and skin conductance, after thorough literature searches in databases like Web of Science, ProQuest, Google Scholar, Baidu Scholar, etc. Thus, it is highly likely that the machine-learning-based multimodality approach for pilot selection is still an unexplored blue sea. No attempts have used machine learning algorithms, heart rate, and skin conductance data to select pilots. Thus, our efforts in vain to locate other similar papers using the multimodality +machine-learning approach may suggest the current study is one of the pioneering works and early adopter of HTC Eye Pro Virtual Reality to select pilots using robust machine learning algorithms and with a more economical and practical platform. Future studies may consider additional machine learning algorithms, especially recent advances in deep learning, such as CNN, RNN, and Transformer algorithms, etc., although this study has extensively compared as many as five algorithms including SVM, KNN, LR, LGBM, and Decision Tree. More complex neural network algorithms were not explored in this current study as accuracy alone is not the only indicator goal for our research; INTERPRETABILITY is another, if not the most important, criterion for pilot selection. More complex algorithms can be attempted with our dataset accumulating year by year. In addition, the pilot selection is not a field similar to "hardware with seldom changes", but more like software that warrants a periodical update. With yearly new data for pilot selection, relatively new concepts in machine learning like "Active Learning" [57] should be considered in future algorithms to update the models yearly to reach an overarching goal of "Faster, Higher, Stronger" algorithms for pilot selection.

## 5. Conclusions

To sum up, this study contributed to pilot selection in at least three aspects: First, our SVM+MIC algorithm achieves a high predictability of 93.33% accuracy, which outperforms most existing SVM, logistic regressions models in literature [32, 33]. Second, the Decision Tree model with 88.89% accuracy shows an interpretable finding that novice pilots are who uses the Altitude Indicator less frequently and have smaller Rudder Input 8s before landing. Third, Virtual Reality with embedded eye tracker and possibility of automated analysis of area of interest, can provide a low-cost, portable, and efficient platform for pilot selection.

**Author Contributions:** For research articles with several authors, a short paragraph specifying their individual contributions must be provided. The following statements should be used "Conceptualization, L.K. and Y.Y.; methodology, L.K.; software, L.K.; validation, L.K., Y.Y. and Z.Z.; formal analysis, L.K.; investigation, L.K.; resources, X.X.; data curation, L.K.; writing—original draft preparation, L.K.; writing—review and editing, L.K.; visualization, L.K.; supervision, X.X.; project administration, L.K.; funding acquisition, Y.Y. All authors have read and agreed to the published version of the manuscript." Please turn to the CRediT taxonomy for the term explanation. Authorship must be limited to those who have contributed substantially to the work reported.